\documentclass[final,5p,times,twocolumn]{elsarticle}

\usepackage{graphicx}
\usepackage{amssymb}
\usepackage{lipsum}
\usepackage{mathtools}
\usepackage{hyperref}
\usepackage{makecell}
\usepackage{subcaption}
\usepackage{graphicx}
\usepackage{caption}
\usepackage{multirow}
\usepackage[all]{xy}
\usepackage{booktabs}
\usepackage[skip=0.1\baselineskip]{caption}
\usepackage[table]{xcolor}
\usepackage{amsmath}

\usepackage{natbib}
\biboptions{semicolon,round}

\setcitestyle{authoryear,open={(},close={)}}

\hypersetup{
     colorlinks   = true,
     citecolor    = blue
}

\journal{NeuroImage}

\begin{document}

\begin{frontmatter}

\title{Ten years of image analysis and machine learning competitions in dementia.}
%\title{The role of image analysis and machine learning competitions in early diagnosis and prediction of dementia.}
% \title{A perspective on grand challenges for classification and prediction in dementia.}
% \title{Grand Challenges on Image-based Quantification, Diagnosis and Prediction in Dementia}

\author[1]{Esther E. Bron}
\ead{e.bron@erasmusmc.nl}
\author[1]{Stefan Klein}
\author[2]{Annika Reinke}
\author[3]{Janne M. Papma}
\author[2]{Lena Maier-Hein}
\author[4]{Daniel C. Alexander}
\author[4]{Neil P. Oxtoby}

\address[1]{Department of Radiology and Nuclear Medicine, Erasmus MC, Rotterdam, the Netherlands}
\address[2]{Division of Computer Assisted Medical Interventions (CAMI), German Cancer Research Center (DKFZ), 69120 Heidelberg, Germany}
\address[3]{Department of Neurology, Erasmus MC, Rotterdam, the Netherlands}
\address[4]{Centre for Medical Image Computing, Department of Computer Science, University College London, London WC1E 6BT, UK}

\begin{abstract}
Machine learning methods exploiting multi-parametric biomarkers, especially based on neuroimaging, have huge potential to improve early diagnosis of dementia and to predict which individuals are at-risk of developing dementia. To benchmark algorithms in the field of machine learning and neuroimaging in dementia and assess their potential for use in clinical practice and clinical trials, seven grand challenges have been organized in the last decade: MIRIAD (2012), Alzheimer’s Disease Big Data DREAM (2014), CADDementia (2014), Machine Learning Challenge (2014), MCI Neuroimaging (2017), TADPOLE (2017), and the Predictive Analytics Competition (2019). Based on two challenge evaluation frameworks, we analyzed how these grand challenges are complementing each other regarding research questions, datasets, validation approaches, results and impact.

The seven grand challenges addressed questions related to screening, \textcolor{black}{clinical status estimation}, prediction and monitoring in (pre-clinical) dementia. There was little overlap in clinical questions, tasks and performance metrics. Whereas this aids providing insight on a broad range of questions, it also limits the validation of results across challenges. The validation process itself was mostly comparable between challenges, using similar methods for ensuring objective comparison, uncertainty estimation and statistical testing. In general, winning algorithms performed rigorous data pre-processing and combined a wide range of input features. 

Despite high state-of-the-art performances, most of the methods evaluated by the challenges are not clinically used. To increase impact, future challenges could pay more attention to statistical analysis of which factors (i.e., features, models) relate to higher performance, to clinical questions beyond Alzheimer's disease, and to using testing data beyond the Alzheimer's Disease Neuroimaging Initiative. Grand challenges would be an ideal venue for assessing the generalizability of algorithm performance to unseen data of other cohorts. Key for increasing impact in this way are larger testing data sizes, which could be reached by sharing algorithms rather than data to exploit data that cannot be shared. Given the potential and lessons learned in the past ten years, we are excited by the prospects of grand challenges in machine learning and neuroimaging for the next ten years and beyond.  

\end{abstract}

%% highlights


\end{frontmatter}



\newcommand\blfootnote[1]{%
  \begingroup
  \renewcommand\thefootnote{}\footnote{#1}%
  \addtocounter{footnote}{-1}%
  \endgroup
}
%\blfootnote{\textit{Abbreviations}: MRI, Magnetic Resonance Imaging; }

\DeclareRobustCommand{\de}[3]{#3}

\newcommand{\beginsupplement}{%
        \setcounter{table}{0}
        \renewcommand{\thetable}{S\arabic{table}}%
        \setcounter{figure}{0}
        \renewcommand{\thefigure}{S\arabic{figure}}%
     }

\section{Introduction}
Over the past decade there has been a great research interest in artificial intelligence approaches assisting the management of dementia. Especially, machine learning methods exploiting multi-parametric biomarkers from large-scale datasets have shown huge potential to detect dementia at an early stage, to provide insight into etiology and to monitor and predict the development of symptoms. Because of this huge potential and increased availability of data, many articles presenting novel algorithms have been published, most of them using neuroimaging as a key biomarker \citep{Falahati2014, Rathore2017, Oxtoby2017, Ansart2021}. 

Since 15 years, grand challenges have been organized in the biomedical image analysis research field. These are international benchmarks in competition form that have the goal of objectively comparing algorithms for a specific task on the same clinically representative data using the same evaluation protocol. In such challenges, the organizers supply reference data and evaluation measures on which researchers can evaluate their algorithms. Over the past years, the number and the impact of such grand challenges has increased \citep{Maier-Hein2018}. Also in the field of machine learning algorithms for dementia, such grand challenges have been organized to gain insight into successful approaches and their potential for use in clinical practice and clinical trials \citep{Cash2015, Allen2016, Bron2015, Sabuncu2014, Sarica2018, Marinescu2018, Fisch2021}.

With the trend of grand challenges in the field of biomedical image analysis becoming more important, two frameworks have been proposed in the literature to evaluate and structure the design of grand challenges and assess the effect of the design on the interpretation and impact of challenge results. First, \cite{Maier-Hein2018} performed an evaluation of 150 challenges and identified major problems, which led to best practice advice. Regarding challenge design, they found that in the majority of challenges it was not possible to reproduce results, to adequately interpret them, and compare them across challenges, since relevant information was often missing and since challenge design (e.g. a choice of metrics and methods for rank computation) was highly heterogeneous. In addition, the authors concluded that the ranking of an algorithm in a challenge was not robust, but highly dependent on a number of design choices such as the test data sets used for validation, the observers who annotated the data and the metrics chosen for performance assessment, as well as the methods used for aggregating values. Second, \cite{Mendrik2019} categorized two types of challenges: challenges that aim to generate insight (insight challenges) and those that aim to solve a problem (deployment challenges). The main purpose of this categorization was to guide researchers in setting-up challenges of which the objectives match with its design and conclusion. Mixing the insight design with the deployment design will negatively impact the clinical significance of the challenge and generalizability of methods. While the first framework is based on a quantitative approach and the latter framework provides a mostly qualitative view on grand challenges, the identified topics of importance for evaluation of challenge design and results largely overlap.

In this review, we will evaluate the grand challenges in dementia based on the topics defined by the frameworks of \cite{Maier-Hein2018} and \cite{Mendrik2019}. Our main research question is how the grand challenges on this topic are strengthening and complementing each other. First, we will highlight unanswered clinical questions in dementia and the related tasks that are being addressed by machine learning algorithms. Subsequently, we will summarize the included grand challenges, with an emphasis on the role of neuroimaging.  We will summarize design choices, including the alignment of their purpose with the relevant questions in dementia, the data and truth criteria, and performance metrics. In addition, we will outline the results and assess their impact. Finally, we will reflect on the status of the field and the open questions that remain to be addressed by future grand challenges. 

\section{Questions in clinical and pre-clinical dementia} 
Worldwide 50 million people are estimated to be living with dementia  and Alzheimer's disease is the most prevalent underlying cause \citep{AlzheimersAssociation2020}. For Alzheimer's disease, the pre-clinical phase is found to take about 20 years before the onset of clinical symptoms \citep{Gordon2018}. Diagnosis of dementia is highly challenging and usually takes a substantial period of time after the first clinical symptoms arise \citep{VanVliet2013}. After diagnosis, novel challenges are related to prediction and monitoring of the disease, both in the natural disease course and under treatment with potential disease-modifying drugs in clinical trials. Along different phases of dementia, clinically relevant questions could be identified to which machine learning methods can contribute by solving related tasks exploiting multi-parametric biomarkers.

First, in the pre-clinical stage, i.e. a community-dwelling population without cognitive complaints, a key question is to identify persons at-risk for developing dementia based on biomarker measurements at an individual level. \textcolor{black}{Several approaches for biomarker-based \textbf{screening} have been proposed \citep{Licher2019, Cole2017, Wang2019}. Regarding neuroimaging, a frequently studied screening biomarker is the brain age gap, i.e. the difference between chronological age and age predicted from MRI using machine learning, which has been shown to identify individuals at risk for dementia in the general population \citep{Wang2019}.} While highly important for improving disease understanding, there is currently no clinical motivation for population screening. As currently no disease-modifying medication could be offered, the benefit for individuals at-risk to identify them before symptoms develop may be very limited. However, while the first Alzheimer's disease drug is entering the market and while the search for disease-modifying treatments continues, screening of individuals at \textcolor{black}{a} very early stage may aid inclusion for clinical trials \citep{Panegyres2016}.

Second, in an early-stage clinical population, i.e. patients with cognitive complaints that visit a memory clinic, the key question is whether the person is developing dementia. The purpose of \textcolor{black}{\textbf{clinical status} estimation is to understand the current ‘condition’ of the patient and to establish a clinical diagnosis that enables} timely decision making regarding care and treatment \citep{prince2011alzheimer}. Relevant \textcolor{black}{components of this clinical status are whether there is a cognitive impairment (e.g. quantified by a memory score, or fulfilling diagnostic criteria such as for} mild cognitive impairment or dementia \citep{McKhann2011, Albert2011}) and what the underlying cause of this cognitive impairment is (e.g. Alzheimer's disease).

In addition to \textcolor{black}{current clinical status}, in this early-stage clinical population, i.e. patients with subjective cognitive decline or a diagnosis of mild cognitive impairment, a second key question is whether symptoms will develop into dementia and in what time frame. The purpose of \textbf{prediction} is \textcolor{black}{to understand the future `condition' of the patient,} mainly to inform treatment and care decisions, but may also guide inclusion of individuals at risk for dementia in clinical trials. Outcomes of interest for prediction are for example diagnosis and clinical functioning at a future time point or time-to-conversion to dementia.

Lastly, in a later-stage clinical population, after diagnosis has been established, a key question involves \textbf{monitoring} the disease development and, if applicable, the patient's response to treatment. In follow-up consultations, biomarkers could be used to assess the disease progression in a quantitative way.

% 2.1 Pre-clinical population - cross-sectional
% Target population: pre-clinical (no memory complaints)
% Question: is this person's brain development abnormal?
% Setting: cross-sectional
% Purpose: to identify targets for prevention
% Task: screening
% Challenges: PAC

% 2.2 Early-stage/clinical population - cross-sectional
% Question: does this person have dementia and which underlying disease
% Task: diagnosis
% Setting: cross-sectional
% Purpose: to make decisions for care and treatment 
% Target population: clinical (subjective memory complaints, mild cognitive impairment, early-stage dementia)
% Challenges: CADDementia, MCI-NI, MLC, DREAM3

% 2.3 Early-stage/clinical population - longitudinal
% Question: is this person at risk of developing dementia?
% Task: prediction
% Setting: longitudinal
% Purpose: inclusion in clinical trials, to make decisions for treatment
% Target population: clinical (subjective memory complaints, mild cognitive impairment)
% Challenges: DREAM1/2, TADPOLE, MCI-NI

% 2.4 Patient population - longitudinal
% Question: how has the disease in this person developed?
% Task: quantification of changes
% Setting: longitudinal
% Purpose: to monitor disease development and to monitor treatment
% Target population: clinical (patients)
% Challenges: MIRIAD

% \begin{figure}[!htb]
% \centering
% \includegraphics[width=1\linewidth]{Timeline.eps}

% \caption[overview]{The seven challenges mapped onto the timeline of clinical questions of screening, diagnosis, prediction and monitoring in dementia.}
% \label{fig:overview}
% \end{figure}

{ % TABLE QUESTIONS
% begin box to localize effect of arraystretch change
\renewcommand{\arraystretch}{2.0}
\begin{table*}[hbt]
\caption{Overview of the tasks for the seven grand challenges in dementia, relating to the clinical problems of screening, \textcolor{black}{clinical status estimation}, prediction and monitoring. AD: Alzheimer's disease, MCI: mild cognitive impairment, CN: cognitively normal. \label{tab:challenge-questions}}
\begin{center}
\begin{footnotesize}
\begin{tabular}{ p{3cm} p{3.5cm} p{3.25cm} p{3.5cm} p{3.25cm}} 
\cellcolor{gray!10}  & \cellcolor{gray!10} \textbf{Screening} & \cellcolor{gray!10} \textbf{\textcolor{black}{Clinical status}} & \cellcolor{gray!10} \textbf{Prediction} & \cellcolor{gray!10} \textbf{Monitoring}  \\
MIRIAD \citep{Cash2015} &  &  &   & \cellcolor{green!25} Estimation of atrophy from MRI.\\
\cellcolor{gray!10}DREAM \citep{Allen2016} & \cellcolor{green!25} Screening for amyloid pathology from genetics, clinical data and demographics (Task 2). & \cellcolor{green!25} Estimation of MMSE from MRI (Task 3).& \cellcolor{green!25} Prediction of MMSE from genetics, clinical data and demographics (Task 1). & \cellcolor{gray!10} \\
CADDementia \citep{Bron2015} &   & \cellcolor{green!25} Classification of AD/MCI/CN from MRI. &  &   \\
\cellcolor{gray!10}MLC\footnotemark[4]   & \cellcolor{green!25} Estimation of age from MRI.   & \cellcolor{gray!10} & \cellcolor{gray!10} & \cellcolor{gray!10} \\
MCI-NI\footnotemark[5] &  & \multicolumn{2}{p{6.75cm + \tabcolsep + \tabcolsep}}{\cellcolor{green!25} Classification of AD/MCI convertor/MCI stable/CN from  MRI.} \\
\cellcolor{gray!10}TADPOLE \citep{Marinescu2018} & \cellcolor{gray!10} & \cellcolor{gray!10} & \cellcolor{green!25} Prediction of AD/MCI/CN, ADAS-Cog13 and ventricle size from multi-parametric biomarkers.  & \cellcolor{gray!10}  \\
PAC \citep{Fisch2021} & \cellcolor{green!25} Estimation of age from MRI. & &  &\\
\end{tabular}
\end{footnotesize}
\end{center}
\end{table*}
}

% (("benchmark study") OR ("grand challenge")) AND ((dementia) OR (alzheimer) OR (AD) OR (vascular) OR (frontotemporal) OR (FTD) OR (DLB) OR (lewy) OR (brain) OR (brainage) OR (aging)) AND ((imaging) OR (MRI) OR (neuroimaging))

\section{Grand challenges in dementia}
Here we discuss grand challenges on screening, \textcolor{black}{clinical status estimation}, prediction and monitoring in (pre-clinical) dementia with a focus on neuroimaging data. \textcolor{black}{Challenges were found by searching challenges listed by \url{grand-challenge.org} and \cite{Maier-Hein2018} as well as searching publications through Pubmed and Google based on key words. Candidate studies were screened by reading the title and if necessary the abstract or website. No criterion on publication year was used. Details regarding the identification and screening can be found in the supplementary information.} Inclusion criteria for studies were 1) being organized as a competition, 2) neuroimaging data included as input, 3) challenge tasks were related to screening, \textcolor{black}{clinical status}, prediction, or monitoring and 4) medical application area was (pre-clinical) dementia. Hence, we excluded benchmarks that were not set up as a grand challenge \citep{cuingnet2011automatic, Wen2020, Ansart2021}, grand challenges that did not include neuroimaging but used for example clock-drawing test data\footnote{\url{aicrowd.com/challenges/addi-alzheimers-detection-challenge}} or speech data \citep{Luz2020}, grand challenges on brain image registration or segmentation \citep{KleinA2009, Mendrik2015}, and grand challenges in diseases other than dementia. 

%We searched grand challenges through a PubMed search, on challenge platforms (Grand Challenge, Kaggle, CodaLab and Synapse), and at conferences (MICCAI, IPMI, OHBM).

%(e.g. the 2019 Alzheimer's Syndrome Prediction challenge \url{http://challenge.xfyun.cn/2019/gamedetail?type=detail/alzheimer},	Alzheimer's Dementia Recognition through Spontaneous Speech \url{http://www.homepages.ed.ac.uk/sluzfil/ADReSS/})

Based on this, we evaluated seven grand challenges:
\begin{itemize}
  \setlength\itemsep{-0em}
  \item Minimal Interval Resonance Imaging in Alzheimer's Disease atrophy challenge (MIRIAD, 2012): the challenge task was to quantify volumes and atrophy rates of several brain structures based on a pair of images blinded for time points with the objective of monitoring dementia \citep{Cash2015}. A total of 9 teams participated.
  %Data consisted of 708 T1-weighted MR images of 69 subjects (46 AD, 23 control), each subject had 1-12 scans acquired during 9 visits in a period of 1-2 years \citep{Malone2013}.
  \item Alzheimer’s Disease Big Data DREAM Challenge (DREAM, 2014)\footnote{\url{synapse.org/##!Synapse:syn2290704}}: the challenge consisted of three tasks: 1) to predict change in cognitive scores based on genotypes imputed from single-nucleotide polymorphism (SNP) array data, 2) to screen for amyloid positivity in cognitively normals based on genotypes, and 3) to estimate the mini-mental state examination (MMSE) score based on T1-weighted MRI \citep{Allen2016}. A total of 32 teams submitted final results (Task 1: 18 teams, 2: 11 teams, 3: 13 teams; all tasks: 4 teams).
  %Training sets consisted of ADNI data (Task 1: N=767, 2: N=176, 3: N-628). Testing data came from the ROS/MAP (Task 1: N=1762, 2: N=257) and AddNeuroMed (Task 3: N=182) datasets. 
  \item Computer-aided diagnosis of dementia challenge (CADDementia, 2014)\footnote{\url{caddementia.grand-challenge.org}}: the challenge task was to classify subjects into the diagnostic categories of Alzheimer's disease, mild cognitive impairment, and cognitively normal controls using a baseline MRI scan \citep{Bron2015}.  Fifteen research teams participated with a total of 29 algorithms.
  %Data consisted of 384 T1-weighted MRI scans, but challenge participants were permitted to use any training data. %Diagnoses were blinded for the test set (N=354).
  \item Machine Learning Challenge (MLC, 2014)\footnote{\url{codalab.org/competitions/1471}}: the challenge consisted of several tasks of binary classification and continuous regression of clinical phenotypes based on T1-weighted MRI. Tasks were blinded to participants. One of the tasks, regression of age in healthy subjects, was included in this evaluation; others were out-of-scope (e.g. schizophrenia diagnosis) \citep{Sabuncu2014}. A total of 12 teams participated.
  \item MCI neuroimaging challenge (MCI-NI, 2017)\footnote{\url{kaggle.com/c/mci-prediction}}: the challenge task was to classify subjects into four clinical categories (Alzheimer's disease, cognitively normal, mild cognitive impairment (MCI), and MCI subjects that converted to Alzheimer's disease) using a baseline MRI scan and MMSE score, thereby combining  \textcolor{black}{current clinical status} and prediction into one task \citep{Sarica2018}. 19 teams participated in the challenge with a total of 347 entries. 
  %Data was randomly selected from the ADNI database (train: N=240, test: N=160). In addition 340 dummy test subjects were created. 
  \item The Alzheimer's Disease Prediction Of Longitudinal Evolution Challenge (TADPOLE, 2017)\footnote{\url{tadpole.grand-challenge.org}}: the challenge consisted of three tasks in predicting future evolution of individuals at risk for Alzheimer's disease  \citep{Marinescu2018, Marinescu2021}. These tasks were to predict three key outcomes: 1) clinical diagnosis, 2) Alzheimer’s Disease Assessment Scale Cognitive Subdomain (ADAS-Cog13), and 3) total volume of the ventricles.  A total of 33 teams participated with 92 algorithms.
  %Training data consisted of 1737 subjects of the ADNI database, but challenge participants were permitted to use any training data. Test data consisted of follow-up data of 219 of these individuals acquired after the challenge deadline.
  \item Predictive Analytics Competition  (PAC, 2019)\footnote{\url{frontiersin.org/research-topics/13501}}: the challenge task was to estimate brain age from healthy individuals (age range: 17-90 years) based on T1-weighted MRI \citep{Fisch2021}. \textcolor{black}{As the literature frequently reported a bias between predicted brain age and chronological age, i.e. overpredicting age of young and underpredicting age of elderly individuals \citep{Treder2021}, a} second task was to estimate brain age with a minimal bias towards chronological age.  A total of 79 teams participated.
  %Training data comprised data from N=2640 subjects and testing was done on N=660 subjects.
  
\end{itemize}

%regarding screening (PAC, DREAM task 2), diagnosis (DREAM task 3, CADDementia, MLC, MCI-NI), prediction (MCI-NI, DREAM task 1, TADPOLE), and monitoring (MIRIAD). The challenges were all unique in their task definition. Input data for most challenges was limited to structural MRI data, with the exception of DREAM and TADPOLE that allowed participants to use multi-parametric data. Output variables were either categorical (CADDementia, MCI-NI, MLC, TADPOLE) or continuous (MIRIAD, DREAM, PAC, MLC, TADPOLE).

% The multi-class classification problem of MCI-NI is a hybrid  betweeen diagnosis and prediction, since of the four classes distiguishing AD/CN/MCI is diagnosis, while MCI into stable and converging groups is prediction.  

\section{Challenge design}
%Challenge design 

%We will assess the problem space of AD classification/prediction (with a focus on the role of neuroimaging) \cite{Mendrik2019}, and the part that is covered by the current challenges (data, truth, metrics).

%Problem
%Purpose statement
%Research questions
%Data 
%Truth criteria
%Metrics

\subsection{Executive summary}
\textcolor{black}{We assessed the design choices made by the seven challenges, as these largely determine the reproducibility, interpretability and comparability of the challenges \citep{Maier-Hein2018}. Table \ref{tab:challenge-questions} summarizes the tasks of all challenges and relates them to the clinical questions. As all challenges had a unique task, different choices have been made regarding data, metrics, and methodology for ranking, which limits comparability.  On the other hand, methods for avoiding cheating, uncertainty estimation and statistical testing are quite similar. An overview of data provided for training and testing can be found in Table \ref{tab:challenge-data}. Table \ref{tab:metrics-truth} provides an overview of truth criteria, anti-cheating strategies, used performance metrics, uncertainty estimates and statistical tests. These design choices, including approaches for challenge and result dissemination, are discussed in more detail in the remainder of this section.}

%How similar are the challenges in terms of metrics? similar metrics for similar tasks? Influence of variation in neuroimaging acquisitions?). %Common standards to strengthen each other. 
%-	How similar are the challenges in terms of rank computation?

\subsection{Research questions and tasks}
The research questions of the seven challenges largely correspond to the four identified clinical questions regarding screening,  \textcolor{black}{clinical status}, prediction and monitoring (Table \ref{tab:challenge-questions}). PAC, MLC and one of the tasks in DREAM (Task 2) address a screening question. PAC and MLC aim to find the most accurate model (and for PAC additionally the most accurate model under a small bias) for estimating brain age in a healthy population, DREAM aims to detect cognitive normals with abnormal amyloid pathology.  \textcolor{black}{Clinical status estimation} is addressed by three challenges in a slightly different way. DREAM estimated the MMSE score from structural MRI (Task 3), while the other two challenges classified different diagnostic groups based on structural MRI (CADDementia: 3 groups, MCI-NI: 4 groups). Prediction is addressed by three challenges that predicted future outcomes in diagnosis (MCI-NI, TADPOLE), cognitive score (DREAM Task 1: MMSE score, TADPOLE: ADAS-Cog13) and MRI measures (TADPOLE: Ventricle volume). In contrary to the  \textcolor{black}{clinical status} challenges, more diverse inputs were used for the prediction tasks (DREAM: genetics, MCI-NI: MRI, TADPOLE: MRI, PET, cognition and cerebrospinal fluid markers). Finally, one challenge addressed monitoring: MIRIAD estimated volume and volume changes from longitudinal structural MRI after diagnosis. All challenges were insight challenges \citep{Mendrik2019} that aim to provide insight into methodological choices for algorithms in the tasks of screening, \textcolor{black}{clinical status}, prediction and monitoring.

%Maier-Hein: Field(s) of application, Task category(ies), Target cohort, Algorithm target(s), Assessment aim(s)

%Assess:
%-	Consistency (of objectives, design and conclusions) will be assessed. 
%-	How were challenges designed, which questions did they as ask and how were these related to an ultimate goal to impact medicine/healthcare, were methods appropriate for the questions asked

\subsection{Data}
Data set sizes differed substantially ranging from 30 to 2640 subjects for training and 69 to 1762 subjects for testing (Table \ref{tab:challenge-data}). Multi-center data was used in all challenges but MIRIAD and MLC. No specific attention was given to diversity of the data. Whereas most challenges used same population training and testing data, DREAM explicitly used data sets of different cohorts. Four challenges allowed participants to use additional training data (DREAM, CADDementia, TADPOLE, PAC). The Alzheimer's Disease Neuroimaging Initiative (ADNI)\footnote{\url{adni.loni.usc.edu}} was used in four challenges: either as (optional) training data (DREAM, CADDementia) or both for training and testing (MCI-NI, TADPOLE). In addition, four challenges had a public leaderboard on their website showing performances on a data subset during the competition phase (DREAM, MLC, MCI-NI, TADPOLE). 

All challenges used a fixed test data set, aiming to compare methods for one specific setting. \textcolor{black}{Labels of the test data were blinded and submission of prediction results was requested.} The test set was in most cases defined in a natural way, e.g. based on visit date (e.g. MIRIAD, TADPOLE) or using a separate cohort (e.g., DREAM, CADDementia). Other challenges used a random split into a training and testing (MLC, MCI-NI, PAC). The challenges could have provided a more complete view of performance in different settings by addressing variance in the challenge design, which can be done by randomizing as many sources of variation as possible and using multiple data splits \citep{Bouthillier2021}.

%Influence of preprocessing/pipelines on TADPOLE results. Makes for an interesting and important discussion paragraph, potentially
Whereas most of the challenges focused on brain MRI as only type of input, TADPOLE and DREAM specifically targetted multi-modal data including cognition, clinical data and genetics. The neuroimaging data was used in different ways; some challenges provided raw imaging data (CADDementia, PAC), whereas other provided pre-computed imaging measurements (MLC, MCI-NI) or allowed for both (DREAM, TADPOLE). Pre-computing image-based features is an advantage for interpretation (differences between methods could be attributed to machine learning methodology only) and participation (more teams can participate, also those without expertise in MRI analysis), but may be a disadvantage for assessing state-of-the-art (variation is limited, promising algorithms may be excluded).

\subsection{Truth criteria and anti-cheating strategies}
To objectify the output of tasks, a reference standard, or truth criterion, should be in place. The used truth criteria by the challenges were clinical diagnosis, follow-up measurements and independent biomarker measurements (Table \ref{tab:metrics-truth}). These criteria do not provide a ground-truth and are all subject to some measurement error. Especially when using clinical diagnosis, it is possible that some of the patients are misdiagnosed as the clinical diagnosis does not take account of post-mortem histopathology or amyloid and tau measurements. None of the challenges reported an analysis of these specific measurement errors or an inter-rater analysis of clinical diagnosis. \textcolor{black}{An alternative approach was used by MIRIAD, which instead of using a direct truth criterion for atrophy quantification used multiple indirect truth criteria focused on variance (i.e., power computation, repeatability, consistency, intra- and intersubject variance).}

To ensure objective comparison, all challenges put some `anti-cheating' strategies in place to promote objective comparison. The most common strategy was to not release test data labels. Using ADNI data for testing however requires an alternative approach as all data is public. TADPOLE therefore required predictions to be submitted before acquisition of the test set by ADNI, and MCI-NI provided fake test data. In this last solution, fake data was not used for the final evaluation but was used for the public leaderboard during the competition phase. As results that include these fake data may have influenced classifier parameters and model selection, final classification performances could have been suboptimal \citep{Castiglioni2018, Donnelly2018}. Another complementary approach, used by all challenges, was to limit the numbers of submissions per team or limit the time between test data release and the submission deadline, in order to reduce the opportunity for model tuning and selection using test set knowledge.

\subsection{Performance metrics}
%Actual performance metrics both the specifics of each challenge and general thoughts on what and how to assess 
For classification tasks, common performance metrics were (balanced) accuracy and area-under-the-ROC-curve (AUC). Accuracy (i.e. correct classification rate) was chosen as primary metric in tasks with balanced data across classes (CADDementia, MCI-NI), while challenges with class imbalance used balanced accuracy or AUC as primary metrics (DREAM, TADPOLE). For regression tasks, common metrics were measures of the differences between values (mean absolute error, root-mean-square error; MLC, TADPOLE, PAC) and correlation (Pearson's or Spearman's; DREAM, MLC). In addition, TADPOLE included the weighted error score and coverage probability accuracy metrics that take account of confidence of the predictions given by the individual methods. MLC asked participants to also submit their 5-fold cross-validation results on the training set in order to analyze the generalization error. The MIRIAD challenge, which had no access to ground truth atrophy measurements, used as a primary metric the required sample size to detect atrophy rate reduction of 25\% with 80\% statistical power, related to potential application of such methods for patient monitoring in clinical trials.

%Which metrics are chosen for ranking the algorithms? Winners (may encourage parameter tuning) or collaborative challenges without winners.How robust are the rankings of algorithms in the individual challenges (Maier-Hein et al., 2018)? 
All grand challenges came up with a final ranking of methods and announced a winner. Ranks were computed on one primary performance metric for each task. Where most challenges ranked methods based a single prediction, DREAM used the median ranking among 100,000 bootstrap resamplings of the test data. TADPOLE summed ranks for their three tasks to make one overall ranking. While such ranking and the possibility to `win' a challenge is highly motivating, a risk of this approach may be in parameter tuning. Instead, collaborative challenges without winners may lead to less parameter tuning and could be more successful in learning about strengths and weaknesses of algorithms \citep{Maier-Hein2018}. 

%Maier-Hein: Metric(s), Justification of metricsa Rank computation method, Interaction level handling, Missing data handling, Uncertainty handling, Statistical tests, Organizer participation policy, Training data policy, Pre-evaluation method, Evaluation software

%-	How is variance addressed \citep{Bouthillier2021}? Account for variance to detect meaningful improvements 
%Whether a statistically significant difference in a metric value is clinically/biologically relevant. 

Most challenges used bootstrapping to estimate confidence intervals on the performance metrics, which is useful for detecting statistically meaningful improvements \citep{Bouthillier2021}. The number of bootstrap samples varied from 50 to 100,000. TADPOLE also evaluated confidence estimations of individual predictions provided by the participants, which is important in clinical application for knowing when estimates are unreliable. In addition, most challenges performed statistical tests, either to test for significance differences between pairs of methods or to test whether performance was significantly better than random guessing. Mostly non-parametric tests were used (permutation testing, Wilcoxon signed-rank), except for CADDementia (McNemar's test).  MCI-NI and PAC did not provide strategies for uncertainty handling and statistical testing. 

\subsection{Dissemination} 
The methods for dissemination differed per challenge. All challenges had a challenge website, mostly connected to a challenge platform (Grand-challenge.org: 2, Kaggle: 1, Codalab: 1, Synapse: 1). Two challenges are still open for new submissions (CADDementia, TADPOLE). For MLC and PAC, websites were not available anymore during our evaluation, so organizers were contacted to provide missing information. Most challenges published an overview paper to present the challenge set-up and discuss results (MIRIAD, CADDementia, TADPOLE, DREAM; TADPOLE additionally published a design paper; CADDementia and DREAM additionally published workshop proceedings with short papers describing individual algorithms). PAC and MCI-NI instead published a special issue with individual papers for the algorithms and an editorial explaining and interpreting the results, and MLC did not publish. Whereas a special issue might be most useful for in-depth understanding and reuse of algorithms, an overview paper may ease results to be interpreted and conclusions to be implemented by the community. We successfully obtained all information required for the overview, which means that aspects determined relevant by \cite{Maier-Hein2018} were reported.

\subsection{\textcolor{black}{ Limitations}} 
All challenges made specific design choices that may have negatively affected their impact, such as using single center data (e.g. MIRIAD), not including the results on all tasks in the publication (e.g. DREAM), using clinical diagnosis as a reference standard (e.g. CADDementia), limiting the use of domain knowledge by blinding the context (e.g. MLC), mixing the clinical questions of  \textcolor{black}{current clinical status} and prediction into one task (e.g. MCI-NI), using ADNI data for both training and testing (e.g. TADPOLE), or not providing strategies for uncertainty handling and statistical testing (e.g. PAC). 
%Reproduction, adequate interpretation, and cross-comparison of results are not possible in the majority of challenges, as only a fraction of the relevant information is reported and challenge design (e.g. a choice of metrics and methods for rank computation) is highly heterogeneous. (Maier-Hein et al., 2018)

\section{Challenge findings}
\subsection{Main findings}
An overview of the main findings and best performing methods can be found in Table \ref{tab:challenge-results}. All challenges aimed to obtain insight into the best strategy for solving the problem and the current state-of-the-art performance. They reported results as performance-based rankings of participating algorithms, compared to the literature and drew conclusions about the contribution of specific modeling choices (i.e. method, features).

There were methodological commonalities and differences across winning methods in the challenges. Winning algorithms generally had special attention for pre-processing of the data and combined a wide range of input features. Also, including more training data or pre-training on other data sets showed advantage. No single modeling methodology stood out among the winners across these challenges, which is perhaps unsurprising given the different objectives of each challenge and the time span involved (10 years) and the plethora of available methods in this field. There was a large variety in types of modeling, varying from Gaussian processes to gradient boosting and from random forests to convolutional neural networks. To illustrate the changes of the field over time: whereas in CADDementia (2014) only one participant used a neural network achieving a relatively low rank, PAC (2019) was dominated by convolutional neural networks achieving high performance.

Winners were announced for all except for two tasks in which algorithms did not outperform random guessing. In the second task of the DREAM challenge, which focused on prediction of cognition based on genetics, participants were unable to develop algorithms with predictive performances significantly better than random. According to \cite{Allen2016}, this might be due to low sample size and trait heterogeneity, but also indicates the difficulty of the task, i.e. information about cognitive resilience is not easily discoverable from genetic analysis. Similarly in the TADPOLE challenge, the ADAS-Cog13 cognitive score was more difficult to forecast than clinical diagnosis or ventricle volume. The only method that performed better than random guessing was a simple mixed effects model. According to \cite{Marinescu2021}, the difficulty could be due to variability in administering the cognitive tests, practice effects and a short follow-up time.

Most challenges looked at qualitative trends in factors \textendash such as data types, features or models\textendash \space leading to higher performance. This evaluation is usually not performed as a statistical analysis. Instead, as earlier concluded by \cite{Ross2021} for image analysis challenges in general, results analysis is often restricted to pure ranking tables, leaving important questions unanswered. For example, quantitative statistical analyses are desirable to properly understand which factors (i.e. features, models) contributed to the ranking, and to find failure modes for any given algorithm, data or objective. Such analysis were not performed by any of the challenges as the challenge design and number of submissions are insufficient.

\subsection{Clinical impact}
Most challenges generated novel insights into the best current strategy for a specific clinical problem related to screening, \textcolor{black}{clinical status}, prediction or monitoring in dementia. Next to conclusions on methodology, unanswered questions were identified (e.g. limited performance in predicting cognitive scores) and some advice for use in practice was given (e.g. use a direct measure of ventricular volume change to assess atrophy in clinical trials). In general, challenges to date in dementia are all insight challenges; their results are not directly applicable in clinical practice but they are rather pushing the state-of-the-art towards answering clinical questions.

Most challenges attempted to distinguish the best performing submissions using statistical significance testing. However, the considerable challenge of determining the practical significance of such statistical differences was generally not tackled  \citep{Mendrik2019}. For example, according to the non-parametric test in TADPOLE, the third-ranked algorithm (multiclass AUC=0.907) performed significantly worse than winning algorithm (multiclass AUC=0.931). It could be argued that such \textcolor{black}{a} difference in performance is not clinically relevant. Indeed, other factors might be more important in such cases, such \textcolor{black}{as} the ease of applying the algorithm and the possibilities for understanding its outputs \citep{Maier-Hein2018}.

Confounding may also impact the results of grand challenges. Awareness of possible confounding is essential to ensure that actual clinical outcomes rather than confounders are modeled. Age is for example a well-known confounder of dementia that should be taken account of \citep{Dukart2011}. This also holds for biases more generally, the presence of biases such as scanner type \citep{Kruggel2010} and comorbidities \citep{Ramirez2016} in challenge data may impact challenge results. For example, despite age-matched classes in the CADDementia test set, challenge outcomes may have been influenced by age as test subjects were younger than the subjects whose data was used by most algorithms for training. 

Key to clinical impact is that the performance of the algorithms reported by the challenges generalizes to real-world clinical data. The dominance of ADNI data in the challenges may limit this generalizability. We observed that all challenges on \textcolor{black}{clinical status} and prediction used ADNI data (2 for training; 2 for training and testing). While the challenges using ADNI for testing took measures to limit the advantage of researchers being more familiar to the data, the challenge results may still be biased towards its specific data characteristics. Data from the ADNI plays a major role in the research field of machine learning in Alzheimer's disease and is used for optimization and validation in $60-90\%$ of the published articles \citep{Rathore2017, Grueso2021}. Therefore, state-of-the-art performance in ADNI is likely to be an overestimation of performance in clinical practice. Grand challenges would be an ideal venue for assessing the generalizability of algorithm performance beyond ADNI to unseen data of other cohorts, ideally real-world clinical data. This was done by the two challenges using ADNI for training resulting in insight in the performance on external test data. However, future challenges and studies should provide insight into which dataset and algorithm characteristics are key to the generalizability of performance. Since in clinical practice there is generally limited training data available that represents new patients, there would be high relevance for future grand challenges to evaluate not only the performance drop but also factors influencing such generalization.

The code of algorithms that participated in challenges are mostly not available to the community for future research. While some individual methods are available through participants' public repositories, none of challenges released algorithms. For the TADPOLE challenge, algorithms were collected post-hoc, resulting in a collection of the code of seven algorithms in a central repository\footnote{\url{tadpole-share.github.io}}. In general, availability of the algorithms, especially those that performed well, would increase the challenge's impact on further research and eventually clinical practice. 

\section{\textcolor{black}{Recommendations for future challenges}}
Most of the methods evaluated by the challenges are not clinically used, which indicates that the field is some way off from making real clinical impact. Based on this review, we identified several key aspects relating to clinical impact in dementia that are not or only partly addressed by current challenges but should, in our opinion, be a primary focus in developing future challenges \textcolor{black}{to increase their contribution to clinical science.}

First, regarding clinical impact, it is key that the tasks being addressed by the algorithms contribute to solving questions relevant to clinicians and also that the algorithms and their output can be understood and used by clinicians. Therefore, future challenges should aim to maximize clinical relevance of the challenge findings for example by maximizing engagement of clinicians in the challenge design. \textcolor{black}{In addition, future challenge should strive for a challenge design that includes evaluation of the understandability of the output by clinicians and evaluation of the interpretability of the algorithms, e.g. using machine learning interpretability techniques \citep{Dyrba2021}.} 

Second, it is key to consider clinical impact beyond Alzheimer's disease and ADNI. Challenges with a wider focus could place the application of algorithms in a broader clinical context by assessing generalizability of performance to real-life applications using clinical or population-based data or by assessing clinical questions such as differential diagnosis, prediction or monitoring in other diseases underlying dementia, e.g. frontotemporal dementia or dementia with Lewy bodies. 

Third, extended statistical evaluation of factors related to higher performance -such as data types, features or models- will contribute to evidence-based development of future methods increasing performance and impact on clinical implementation \citep{Ross2021}. However, such a statistical analysis requires a test set of sufficient sample size and a sufficient number of challenge submissions. 

\textcolor{black}{Finally,} a key factor complicating the organization of challenges with either large sample sizes or with clinical questions beyond Alzheimer's disease and ADNI is the collection of clinical test data that can be shared with participants. Developments in research software and infrastructure may provide a solution: sharing algorithms rather than the data.  While already done by some challenges\footnote{\url{mrbrains18.isi.uu.nl}}\textsuperscript{,}\footnote{\url{node21.grand-challenge.org}}, dementia challenges have not yet used this approach. Such an approach would only share a small training dataset, but would keep test data private. Wrapping an algorithm in a container (e.g. Docker\footnote{\url{www.docker.com}}, Singularity \citep{Kurtzer2017}) and applying the algorithms locally to the data (at one site or multiple sites in a federated approach) enables challenges to benchmark algorithms on large sets of data that cannot leave their respective institutes. \textcolor{black}{This could improve both data set size and data set diversity.} Such an approach could be also used for enabling training on larger datasets (i.e. federated learning). The nature of dementia neuroimaging data needing a lot of data-preprocessing may make this more difficult. Platforms such as \url{grand-challenge.org} provide an infrastructure that can be used by challenges without having to implement a custom evaluation system. 

\textcolor{black}{Future challenges should strive to have real impact on dementia science by addressing these recommendations. In short, tomorrow's ideal dementia challenge will: 1) encourage explainability and interpretability; 2) address key clinical questions such as early detection, non-AD dementia, and differential diagnosis; 3) include a thorough statistical evaluation to map out the performance landscape; and 4) strive for unbiased generalizability such as by using privacy-preserving technologies.}

\section{Conclusion}
We evaluated grand challenges benchmarking algorithms in the field of screening,  \textcolor{black}{clinical status}, prediction and monitoring of dementia based on neuroimaging and additional data. We used the frameworks of \cite{Maier-Hein2018} and \cite{Mendrik2019} to identify topics for this evaluation. 

Our main research question was how the grand challenges in dementia strengthened and complemented each other. Although the number of challenges was small (n=7), they had little overlap: they addressed different clinical questions, had unique tasks and made different choices regarding data, metrics and methodology for ranking. Methods used for avoiding cheating, uncertainty estimation and statistical testing were similar. Together the challenges provide valuable insight into the state-of-the-art and identified where key limitations currently exist. In general, winning algorithms made an effort regarding data pre-processing and combined a wide range of input features. Whereas complementarity is a strength providing insight on a broad range of questions, it also limits the validation of results between challenges.

As dementia continues to become increasingly important for 21st-century medicine, we are encouraged by the lessons learned from the past ten years of machine learning and neuroimaging grand challenges. This review of current challenge frameworks in dementia showed that challenges have been highly complementary and provided new insights. Future challenges could increase impact \textcolor{black}{by evaluation of interpretability and understandability by clinicians,} by more detailed evaluation of factors related to performance, and by addressing the performance of algorithms for a wider range of clinical questions and for a larger variety of test data. We are excited by the prospects for the next ten years and beyond.

\section*{Acknowledgements}
E.E. Bron acknowledges support from Medical Delta (Diagnostics 3.0: Dementia and Stroke), Dutch Heart Foundation (PPP Allowance, 2018B011), the Dutch CardioVascular Alliance (Heart-Brain Connection: CVON2012-06, CVON2018-28) and the Netherlands eScience Center (2018 Young eScientist award). N.P. Oxtoby is a UKRI Future Leaders Fellow (MR/S03546X/1). N.P. Oxtoby and D.C. Alexander acknowledge funding from the E-DADS project (EU JPND), and the National Institute for Health Research University College London Hospitals Biomedical Research Centre. This work is part of the EuroPOND initiative, which is funded by the European Union's Horizon 2020 research and innovation program under grant agreement No. 666992. Part of this work was funded by the Helmholtz Imaging Platform (HIP), a platform of the Helmholtz Incubator on Information and Data Science. We further acknowledge the Deep Dementia Phenotyping (DEMON) network\footnote{\url{demondementia.com}}, an international network for the application of data science and AI to dementia research.

%\bibliographystyle{elsarticle-num-names} 
%\bibliography{adniparelsnoer.bib}

%%Harvard
\bibliographystyle{model2-names}\biboptions{authoryear}
\bibliography{dementia_challenge_review.bib}

%\clearpage

%\section{Supplementary Material}
%\beginsupplement

{ % TABLE DATA 
% begin box to localize effect of arraystretch change
\renewcommand{\arraystretch}{2.0}
\begin{table*}[htb]
\caption{Overview of data for the seven challenges. Leaderboard data was used to compute algorithm performance for a public leaderboard on the challenge website during the competition phase.\label{tab:challenge-data}}
\begin{center}
\begin{footnotesize}
\begin{tabular}{ p{2cm} p{3.5cm} p{3.5cm} p{3.5cm} p{3.5cm}} 
\cellcolor{gray!10}  & \cellcolor{gray!10}\textbf{Training data} & \cellcolor{gray!10}\textbf{Extra training data} & \cellcolor{gray!10}\textbf{Leaderboard data} & \cellcolor{gray!10}\textbf{Test data}\\
MIRIAD \citep{Cash2015} & 69 subjects (46 Alzheimer's disease, 23 controls), 1-12 scans in 9 visits during 1-2 years. 708 images. \citep{Malone2013} & N.A.  & N.A.  & Same as training data.\\
\cellcolor{gray!10}DREAM \citep{Allen2016}& \cellcolor{gray!10}ADNI data: clinical data, imputed genotypes, (processed) MRI data   (Task 1:  N=767,  2:  N=176,  3: N=628). & \cellcolor{gray!10} AddNeuroMed: clinical data, imputed genotypes (Task 1: N=409) & \cellcolor{gray!10} A subset of the test data (Task 1: N=588, 2: N=129, 3: N=94). Leaderboard usage was limited (Task 1: 100x, Task 2 and 3: 50x). & \cellcolor{gray!10} ROS/MAP: clinical data, imputed genotypes (Task 1: N=1762, 2: N=257) and AddNeuroMed: clinical data, imputed genotypes, (processed) MRI data (Task 3: N=182)\\
CADDementia \citep{Bron2015} & CADDementia data from Erasmus MC, VU University Medical Center and University of Porto: MRI + diagnostic labels (N=30)  &   Any training data permitted.  &  N.A. & CADDementia data: MRI (N=354). Data was roughly balanced over the classes. \\
\cellcolor{gray!10}MLC\footnotemark[4]  & \cellcolor{gray!10} OASIS data \citep{Sabuncu2014}: pre-processed MRI + label (N=315, split into a training and testing set). & \cellcolor{gray!10} N.A. & \cellcolor{gray!10} All test data. & \cellcolor{gray!10} OASIS data: pre-processed MRI, split from training data, size unknown. \\
MCI-NI\footnotemark[5]  & ADNI data: randomly selected, pre-processed MRI, MMSE score and classification label (N=240). & N.A. & Approximately half of the test data (including dummy data). & ADNI (N=160), dummy data (N=340). Data was balanced over the classes. Dummy data was excluded for the final evaluation. \\
\cellcolor{gray!10}TADPOLE \citep{Marinescu2018} & \cellcolor{gray!10} ADNI data: multi-parametric including (pre-processed) structural MRI, diffusion MRI, PET imaging, cerebrospinal fluid biomarkers and cognitive tests (N=1737) & \cellcolor{gray!10}Any training data permitted. & \cellcolor{gray!10} A subset of the training data. & \cellcolor{gray!10} ADNI data: follow-up diagnosis, ADAS-Cog13 score and ventricle volume (N=219).  \\
PAC \citep{Fisch2021} & Data from Imperial College London \citep{Cole2017} and the Institute of Translational Psychiatry Münster: MRI + age (N=2640) & Any training data permitted. & N.A. & Data from Imperial College London \citep{Cole2017} and the Institute of Translational Psychiatry Münster: MRI (N=660)\\
\end{tabular}
\end{footnotesize}
\end{center}
\end{table*}
}

{ % TABLE METRICS
% begin box to localize effect of arraystretch change
\renewcommand{\arraystretch}{2.0}
\begin{table*}[hbt]
\caption{Overview of performance metrics and truth criteria for the seven challenges. AD: Alzheimer's disease, MCI: mild cognitive impairment, MMSE: mini-mental state examination, PET: positron emission tomography. \label{tab:metrics-truth}}
\begin{center}
\begin{footnotesize}
\begin{tabular}{ p{2cm} p{2cm} p{2cm} p{2.5cm} p{2.5cm} p{2cm}  p{2.5cm}} 
\cellcolor{gray!10} & \cellcolor{gray!10} Task & \cellcolor{gray!10} \textbf{Truth criteria} & \cellcolor{gray!10}\textbf{Anti-cheating} & \cellcolor{gray!10} \textbf{Performance metrics} & \cellcolor{gray!10} \textbf{Uncertainty est.} &  \cellcolor{gray!10} \textbf{Statistical tests} \\
MIRIAD \citep{Cash2015} & Estimation of atrophy. & N.A., instead reliability and variance of atrophy measurements were assessed \citep{Fox2011}. & Timepoint labels of scans blinded (only one baseline labeled) & Sample size to detect a 25\% atrophy rate reduction at 80\% power. Secondary: repeatability, consistency, inter- and intra-subject variance. & Confidence intervals from 2000 bootstrap samples & Non-parametric tests between pairs of methods (2000 bootstrap samples, $\alpha<0.05$)\\
\cellcolor{gray!10}DREAM \citep{Allen2016} & \cellcolor{gray!10}Prediction of cognition (Task 1) &\cellcolor{gray!10}
MMSE score established after 2 years. & \cellcolor{gray!10}Test data outcomes were not released. A maximum of 2 entries per task per team. & \cellcolor{gray!10} Pearson's correlation, Spearman's correlation (using clinical input only and a combination of clinical and genetics input) & \cellcolor{gray!10}Confidence intervals from 100,000 bootstrap samples. & \cellcolor{gray!10}Non-parametric tests (10,000 permutations) to assess if methods performed better than random ($\alpha<0.05$). \\
& Screening of amyloid pathology (Task 2) & Amyloid positivity from PET or cerebrospinal fluid. & ''  & Balanced accuracy, area-under-the-curve & '' & '' \\
\cellcolor{gray!10}& \cellcolor{gray!10}Estimation of cognition (Task 3).  & \cellcolor{gray!10}MMSE score. & \cellcolor{gray!10}''  & \cellcolor{gray!10}Pearson’s correlation, Lin’s concordance correlation coefficient & \cellcolor{gray!10}'' & \cellcolor{gray!10}'' \\
CADDementia \citep{Bron2015} &  Classification of AD, MCI, and cognitively normals.  &  Clinical diagnosis by multi-disciplinary consensus. & Test data labels not released. A maximum of 5 entries per team. & Accuracy. Secondary: Area-under-the-curve, True positive fraction & Confidence intervals from 1000 bootstrap samples &  McNemar's test between pairs of methods ($\alpha<0.05$). \\
\cellcolor{gray!10}MLC\footnotemark[4]   & \cellcolor{gray!10}Estimation of age. & \cellcolor{gray!10}Chronological age. & \cellcolor{gray!10}Tasks and labels blinded, test labels not released. A maximum of 15 entries per team.  & \cellcolor{gray!10}Root-mean-square error, Pearson's correlation. Secondary: Generalization of performance w.r.t. 5-fold cross-validation. & \cellcolor{gray!10}Confidence intervals from 10,000 bootstrap samples. & \cellcolor{gray!10}N.A. \\
MCI-NI\footnotemark[5] & Classification of AD, MCI converting to AD, stable MCI, and cognitively normals. & Diagnosis using clinical criteria and follow-up diagnosis after 48 months for MCI patients. & Public and private leaderboard, but not final test data, were inflated by dummy data. A maximum of 1 entry per team. & Accuracy & N.A. & N.A. \\
\cellcolor{gray!10}TADPOLE \citep{Marinescu2021} & \cellcolor{gray!10}Prediction of a future status of AD, MCI, and cognitively normal (Task 1) & \cellcolor{gray!10}Diagnosis established at a follow-up time point.  & \cellcolor{gray!10} Evaluation data acquired after challenge deadline. A maximum of 3 entries per team. & \cellcolor{gray!10}Area-under-the-curve. Secondary: Balanced classification accuracy & \cellcolor{gray!10}Confidence intervals from 50 bootstrap samples & \cellcolor{gray!10} Non-parametric test between pairs of methods (50 bootstrap samples of the test set, $\alpha<0.05$).\\
& Prediction of ADAS-Cog13 and ventricle size (Task 2\&3) & Variables established at a follow-up time point. &  '' & Mean absolute error. Secondary: weighted error score, coverage probability accuracy & '' & Wilcoxon signed-rank test between pairs of methods ($\alpha<0.05$). \\
\cellcolor{gray!10}PAC \citep{Fisch2021} & \cellcolor{gray!10}Estimation of age & \cellcolor{gray!10}Chronological age. & \cellcolor{gray!10}Age for testing data not released. Limited time (1 week) between test data release and submission deadline. & \cellcolor{gray!10}Mean absolute error. Secondary: Bias using Spearman's correlation of predicted age difference and chronological age.  & \cellcolor{gray!10}N.A. & \cellcolor{gray!10}N.A. \\
\end{tabular}
\end{footnotesize}
\end{center}
\end{table*}
}

{ % TABLE Results
% begin box to localize effect of arraystretch change
\renewcommand{\arraystretch}{2.0}
\begin{table*}[htb]
\caption{Overview of the main findings of the seven challenges.\label{tab:challenge-results}}
\begin{center}
\begin{footnotesize}
\begin{tabular}{ p{2cm} p{4.5cm} p{4.5cm} p{4.5cm} } 
\cellcolor{gray!10}  & \cellcolor{gray!10}\textbf{Main research question}  & \cellcolor{gray!10}\textbf{Main findings} & \cellcolor{gray!10}\textbf{Best methods}\\
MIRIAD \citep{Cash2015} & What is the best current strategy for atrophy measurement from MRI and what is the required sample size for hypothetical clinical trials using such methods? & Measures of volume change, particularly ventricle and whole brain, were consistent and robust, leading to stable sample size estimates. Hippocampal measures were more variable, likely due to the differing definitions of the structure. Direct measures of change had smaller variance than indirect measures. & The methods achieving the smallest required sample sizes were the Boundary Shift Integral (for whole brain atrophy measurement) and the combination of Demons-LCC registration and regional flux analysis (for ventricle and hippocampus atrophy measurement).\\

\cellcolor{gray!10}DREAM \citep{Allen2016}& \cellcolor{gray!10} What are the current capabilities for estimation of cognition and prediction of cognitive decline using genetic and imaging data from public data resources? & \cellcolor{gray!10}Predictions of cognitive performance developed from genetic or structural imaging data were modest across a diverse set of models, with most algorithms performing roughly equivalently. The data was probably inadequate to support these tasks. There was no clinical value in prediction of cognitive decline based on genetic data. & \cellcolor{gray!10}The algorithm that generated the best absolute mean combined rank in task 1 and 3 used Gaussian process regression. For task 2, no method performed better than random.\\

CADDementia \citep{Bron2015} & What is the best current strategy for classification of Alzheimer's disease and mild cognitive impairment based on MRI and how does it perform in a clinically representative multi-center data set? &  The best performing algorithms incorporated features describing different properties of the scans such as shape and intensity features in addition to volume and thickness. Performance was additionally influenced by the classification model and training data size. While all methods outperformed random guessing, performance was too low for clinical application. & The method that achieved best accuracy used linear discriminant analysis of a combination of features measuring volume, thickness, shape and intensity relations of brain regions.\\

\cellcolor{gray!10}MLC\footnotemark[4]  &  \cellcolor{gray!10} What is the current state of the art in the field of neuroimage-based prediction? & \cellcolor{gray!10}\textit{An interpretation of the findings has not been published.} & \cellcolor{gray!10}\textit{Descriptions of individual methods are published.}\\

MCI-NI\footnotemark[5] & What is the best current strategy for prediction based on structural MRI features of early diagnosis of Alzheimer's disease, differential diagnosis of mild cognitive impairment and conversion into Alzheimer's disease? & \textit{Articles on individual algorithms were published\footnote{\url{sciencedirect.com/journal/journal-of-neuroscience-methods/vol/302}}, but an overall interpretation of the findings has not been published.} & The method that achieved  best accuracy used an ensemble of random forest classifiers and morphological MRI features. \\
\cellcolor{gray!10}TADPOLE \citep{Marinescu2021} & \cellcolor{gray!10} What are the data, features and approaches that are the most predictive of future progression of subjects at risk of Alzheimer's disease to aid identification of patients for inclusion in clinical trials who are in early stages of disease and are likely to progress over the short-to-medium term (1-5 years)? & \cellcolor{gray!10} ADAS-Cog13 scores were more difficult to forecast than clinical diagnosis or ventricle volume. No single method performed well on all prediction tasks, while  ensemble methods yielded consistently strong results. Diffusion MRI and cerebrospinal fluid biomarkers are most associated with high prediction performance of clinical diagnosis. & \cellcolor{gray!10}The method that achieved the overall best rank used a gradient boosting regression method based on a combination cognitive, clinical and imaging data.\\
PAC \citep{Fisch2021} & What is the best current strategy for accurate prediction of age from brain MRI in an healthy population and what is the best strategy while avoiding bias of age estimation towards the mean age of the training dataset? & Deep learning models performed better than classic machine learning algorithms. & An ensemble of lightweight convolutional neural networks pretrained on UK Biobank data achieved the best performance (MAE=2.90 years, which slightly increased after bias correction to MAE=2.95 years.) \\
\end{tabular}
\end{footnotesize}
\end{center}
\end{table*}
}

\end{document}